\documentclass[a4paper,twoside]{article}

\usepackage{epsfig}
\usepackage{subcaption}
\usepackage{calc}
\usepackage{amssymb}
\usepackage{amstext}
\usepackage{amsmath}
\usepackage{amsthm}
\usepackage{multicol}
\usepackage{pslatex}
\usepackage{apalike}
\usepackage{algorithm2e}
\usepackage[bottom]{footmisc}

\usepackage{graphicx}
\usepackage[hyphens]{url}
\usepackage[most]{tcolorbox}
\usepackage{xcolor}
\usepackage{colortbl}
\usepackage{xfp}
\usepackage{enumitem}
\usepackage{multirow}

\usepackage{SCITEPRESS}     
\usepackage{hyperref}       
\hypersetup{colorlinks=true, urlcolor=blue, citecolor=black, linkcolor=black, breaklinks=true}

\newcommand{\maeheat}[1]{%
  \edef\HeatValue{\fpeval{round(max(8,min(55,8 + 47*(2.4-(#1))/(2.4-0.5))),0)}}%
  \edef\HeatColorArg{blue!\HeatValue}%
  \expandafter\cellcolor\expandafter{\HeatColorArg}#1%
}

\newcommand{\maeheatbf}[1]{%
  \edef\HeatValue{\fpeval{round(max(8,min(55,8 + 47*(2.4-(#1))/(2.4-0.5))),0)}}%
  \edef\HeatColorArg{blue!\HeatValue}%
  \expandafter\cellcolor\expandafter{\HeatColorArg}\textbf{#1}%
}

\newcommand{\meaheat}[1]{%
  \edef\HeatValue{\fpeval{round(max(8,min(55,8 + 47*((#1)-0.0)/(0.61-0.0))),0)}}%
  \edef\HeatColorArg{green!\HeatValue}%
  \expandafter\cellcolor\expandafter{\HeatColorArg}#1%
}

\newcommand{\meaheatbf}[1]{%
  \edef\HeatValue{\fpeval{round(max(8,min(55,8 + 47*((#1)-0.0)/(0.61-0.0))),0)}}%
  \edef\HeatColorArg{green!\HeatValue}%
  \expandafter\cellcolor\expandafter{\HeatColorArg}\textbf{#1}%
}

\begin{document}

\title{Simulating Tenant Responses to Energy Policy Interventions with Transaction-Cost-Aware LLM Agents}

\author{\authorname{Weijie Xia\sup{1}, Stefanie Horian\sup{1}, Hanyue Huang\sup{2}, Queena K. Qian\sup{1}, Jie Yang\sup{1} and Pedro P. Vergara\sup{1}}
\affiliation{\sup{1}Delft University of Technology, Delft, The Netherlands}
\affiliation{\sup{2}Technical University of Munich, Munich, Germany}
\email{w.xia@tudelft.nl, s.horian@tudelft.nl, k.qian@tudelft.nl, J.Yang-3@tudelft.nl, p.p.vergarabarrios@tudelft.nl, hanyue.huang@tum.de}
}


\abstract{Recent studies use Large language models (LLMs) to simulate human opinions and decisions by prompting models with demographic, attitudinal, or persona-based descriptions. Yet such simulations rarely model the practical, cognitive, or social frictions that shape how people respond to policy interventions. Perceived transaction cost (PTC) provides a useful lens for modeling the practical frictions that shape policy responses, such as information burden, administrative effort, coordination demands, and perceived uncertainty. We use this lens to develop a friction-aware persona modeling approach for LLM-based simulation. In the context of energy-efficient renovation (EER), tenants are represented not only by who they are demographically, but by how they perceive the costs, benefits, barriers, and uncertainties associated with proposed renovation plans. Using survey data collected from 1,068 tenants in the Netherlands, comprising approximately 40,548 survey question and answer pairs, we compare prompt-only and fine-tuned settings across GPT-3.5-turbo, Ministral-8B-Instruct, and Llama-3.1-8B-Instruct, and evaluate supervised fine-tuning (SFT) and Group Relative Policy Optimization (GRPO) for local open-weight models. Results show that incorporating PTC-based personas and reasoning consistently improves model performance across both prompt-only and fine-tuned settings, suggesting that PTC-based persona design provides a useful bridge between institutional policy theory and interpretable LLM-based policy simulation.\thanks{Code and data are available at: \href{https://github.com/xiaweijie1996/socialagent}{Personal Repo} and \href{https://github.com/distributionnetworksTUDelft/LLMAgentEnergyCitizent}{TU Delft Repo}.}}

\onecolumn \maketitle \normalsize \setcounter{footnote}{0} \vfill

\section{\uppercase{Introduction}}
\label{sec:introduction}

\noindent The effectiveness of energy policy interventions depends not only on tenants'
preferences, but also on their ability to act on those preferences
\cite{liu2023willingness}. Energy-efficient renovation (EER), for example, may
require tenants to understand proposed measures, assess financial consequences,
coordinate with landlords or contractors, tolerate disruption, and evaluate
uncertain future benefits. These informational, administrative, and
coordination burdens can shape policy responses even when tenants support the
policy's underlying objective \cite{li2025forging}. Policymakers therefore need
tools that represent both tenants' attitudes and the practical frictions
affecting their decisions.

Large language models (LLMs) offer a possible basis for such tools because they
can generate context-sensitive responses from textual descriptions of people
and policy settings. Previous studies have used LLMs as simulated survey
respondents, economic agents, and interactive social agents
\cite{argyle2023out,aher2023using,horton2023large,park2023generative}. Although
LLMs can reproduce some aggregate patterns in human responses, their outputs
may be biased, insufficiently variable, prompt-sensitive, and unreliable for
subgroup inference \cite{bisbee2024synthetic,qu2024performance}. Evidence from
climate and energy research further indicates that demographic conditioning
alone is often insufficient: simulation fidelity improves when prompts include
issue-specific attitudes and covariates \cite{lee2024globalwarming,fell2024energy}.
However, the action-related frictions through which citizens evaluate and
respond to policy interventions remain underrepresented in LLM persona design.

Transaction cost (TC) theory originally emphasized the costs of searching
for information, negotiating, coordinating, monitoring, and implementing
exchanges \cite{coase1937nature,williamson1981economics,north1990institutions},
and environmental policy research similarly shows that such costs shape
policy design and performance \cite{mccann2005transaction,mccann2013transaction}.
In household energy decisions, relevant frictions include time and cognitive
effort, procedural burden, disruption, distrust, and uncertainty
\cite{mundaca2007transaction,lundmark2024understanding}. We focus on
tenants' perceptions of these frictions alongside perceived policy benefits,
such as comfort, health, and energy savings, so that the framework captures
both obstacles to action and motivations for acting. To this end, we adopt
perceived transaction costs (PTCs) as a theoretically grounded way to
represent these frictions.

Motivated by the gap between PTC and current LLM-based persona design, we propose a PTC-aware LLM framework for simulating tenant responses to EER policies. Rather than relying on demographic priors alone, the framework represents tenants through empirically derived profiles of perceived barriers and benefits, and its main contribution is to introduce PTC-aware persona modeling as a bridge between institutional policy theory and LLM agent design. We operationalize this framework with survey data from 1,068 respondents in the Netherlands, evaluating a prompt-only GPT-3.5-turbo baseline alongside two fine-tuned open-weight models, Ministral-8B-Instruct \cite{mistralai2024ministral8b} and Llama-3.1-8B-Instruct \cite{ollama2024llama31}, adapted with QLoRA under supervised fine-tuning (SFT) and Group Relative Policy Optimization (GRPO) \cite{shao2024deepseekmath}. Overall, our results show that grounding LLM personas in PTC consistently improves simulation accuracy over demographic-only prompting, providing a concrete bridge between institutional policy theory and interpretable LLM-based policy simulation.

\section{\uppercase{Literature Review}}

\subsection{LLM-Based Social and Citizen Simulation}
\noindent LLMs have recently been proposed as tools for computational social science because they can classify, explain, and generate social data in ways that complement traditional survey and annotation pipelines \cite{ziems2024can}. One line of work uses LLMs to simulate individual or group behavior. Aher, Arriaga, and Kalai introduce ``Turing Experiments'' to test whether LLMs can replicate human-subject experiments across economic, psycholinguistic, and social-psychological settings \cite{aher2023using}. Horton and colleagues frame LLMs as simulated economic agents that can be assigned endowments, preferences, and information before being placed in experimental scenarios \cite{horton2023large}. In human-computer interaction, Social Simulacra and Generative Agents show how LLM-driven agents can populate social environments, remember experiences, plan, and generate plausible interactions \cite{park2022social,park2023generative}.

For public-opinion research, Argyle et al. introduce the idea of conditioning LLMs on demographic backstories from real survey respondents and evaluating whether the resulting ``silicon samples'' reproduce human response patterns \cite{argyle2023out}. This work motivates survey-conditioned simulation, but subsequent studies caution against treating synthetic responses as direct substitutes for human survey data. Bisbee et al. find that LLM-generated survey responses can match broad averages while producing too little variation, unstable results across prompt changes, and regression patterns that differ from human surveys \cite{bisbee2024synthetic}. Qu and Wang similarly document performance variation and demographic bias when simulating public opinion across countries and topics \cite{qu2024performance}. Together, these studies suggest that LLM-based citizen simulation should be empirically benchmarked, context-specific, and explicit about its conditioning variables.

\subsection{Persona Modeling and Survey Conditioning}
\noindent Persona design is central to LLM simulation because the model's response is shaped by the attributes and context provided in the prompt. Existing work commonly conditions personas on demographic attributes, political identity, psychographic traits, prior survey answers, or task-specific covariates \cite{argyle2023out,lee2024globalwarming}. In climate-opinion simulation, Lee et al. show that demographic-only prompts can fail to capture global-warming beliefs, while adding issue-relevant covariates such as involvement, interpersonal discussion, and perceived scientific consensus improves fidelity \cite{lee2024globalwarming}. Fell's energy-social-survey replications likewise demonstrate the promise of population-representative LLM agents for energy research while emphasizing practical and ethical limitations \cite{fell2024energy}.

These findings motivate persona representations that go beyond demographic identity alone. For energy policy interventions, what matters is not only who a tenant is, but also how difficult it is for that tenant to act. Consider two otherwise comparable respondents evaluating the same home-renovation subsidy. One can quickly find reliable information, compare installers, understand eligibility rules, assemble the required documents, and coordinate the work with contractors. The other struggles to identify trustworthy advice, interpret administrative requirements, estimate the likely benefits, or align decisions with landlords, family members, or service providers. These differences reflect TCs such as information search, administrative burden, coordination demands, and uncertainty, all of which can directly mediate policy response \cite{mundaca2007transaction,lundmark2024understanding}. A PTC-based persona therefore provides theoretically grounded context about the mechanisms through which tenants translate policy offers into action. This approach differs from generic role prompting because the persona dimensions are derived from TC theory and energy policy evidence rather than from intuitive or ad hoc descriptions of tenants.

\subsection{Transition Cost in Social Science and Policy}
\noindent TC begins from the observation that economic exchange and institutional coordination are not frictionless. Coase's account of the firm explains organizational boundaries through the costs of using the price mechanism \cite{coase1937nature}. Williamson develops TC economics around the transaction as the unit of analysis, emphasizing uncertainty, asset specificity, bounded rationality, opportunism, and governance structures \cite{williamson1981economics}. North extends this logic to institutions, arguing that formal and informal rules shape human interaction partly by structuring transaction and production costs \cite{north1990institutions}.

In public and environmental policy, TC  affect not only firms but also agencies, intermediaries, and citizens. McCann et al. argue that policy choice and policy design should account for TC and provide guidance for measuring them in environmental and natural-resource policies \cite{mccann2005transaction}. McCann later synthesizes empirical evidence to show that TC interact with policy design, property rights, institutional settings, and abatement costs \cite{mccann2013transaction}. For energy efficiency, Mundaca shows that tradable white-certificate schemes create costs related to information search, customer persuasion, negotiation, measurement, and verification \cite{mundaca2007transaction}. Lundmark's study of Swedish residential energy renovations estimates substantial household TC and connects them to uncertainty, cognitive limitations, social connectedness, and implementation frictions \cite{lundmark2024understanding}.

\begin{figure*}[t]
\centering
\includegraphics[width=\textwidth,height=1\textheight,keepaspectratio]{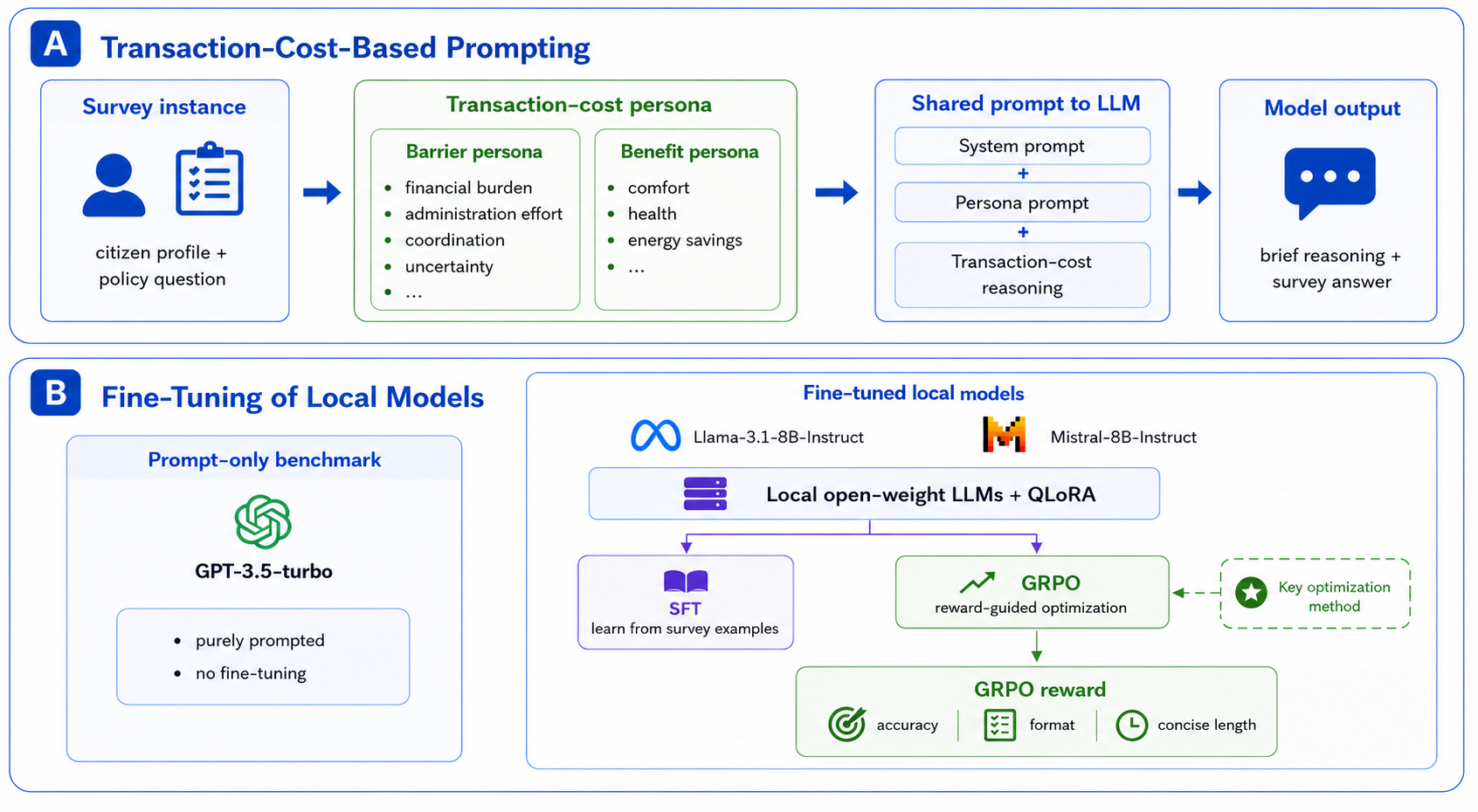}
\caption{Overview of the proposed framework for simulating tenant responses to energy policy interventions.}
\label{fig:method-overview}
\end{figure*}

\section{\uppercase{Method}}

\begin{figure}[!t]
\centering
\begin{tcolorbox}[
    colback=blue!12,
    colframe=blue!35!black,
    arc=4mm,
    boxrule=0.6pt,
    left=2mm,
    right=2mm,
    top=2mm,
    bottom=2mm,
    width=0.98\linewidth
]

\footnotesize

\noindent \textbf{System Prompting:}\\
You answer survey questions for one specific resident. Stay consistent with that person's housing situation, attitudes, and priorities.

\vspace{0.8em}

\noindent \textbf{PTC Persona Prompting:}\\
\textit{Financial Sensitive:} This person generally supports energy-efficient renovation but worries about money, especially higher rent or service charges. Disruption matters less than financial risk. They need clear reassurance that renovation will not make their finances worse.

\vspace{0.6em}

\noindent \textit{Immediate Utility Seekers:} This person mainly cares about direct personal gains, such as better comfort, health, well-being, and lower energy use. Appearance or neighborhood effects matter less. They respond well when short-term benefits are clear.

\vspace{0.8em}

\noindent \textbf{PTC Reasoning Prompting:}\\
Before the final response, briefly think about:
\begin{enumerate}[leftmargin=1.5em, itemsep=0pt, topsep=0.2em]
    \item Burden: time, effort, disruption, moving, or financial risk
    \item Uncertainty: whether the effects are clear or unclear
    \item Personal gains: comfort, health, well-being, and energy savings
    \item Likely direction: stay neutral when information is unclear; lean positive when direct benefits are clear; lean negative when burdens or costs are clear
\end{enumerate}

\vspace{0.8em}

\noindent \textbf{Question:}\\
Please indicate to what extent you personally care about the following benefit of Energy-Efficient Renovation: Increasing thermal comfort for my home (e.g., warmer in winter and cooler in summer).\\
1--I don't care at all; 2--I don't care; 3--Neutral; 4--I care; 5--I care a lot

\vspace{0.8em}

\noindent \textbf{Output Format:}\\
\textbf{\textless thinking\textgreater}\\
 brief reasoning in 2--4 sentences\\
\textbf{\textless /thinking\textgreater}\\
\textbf{\textless answer\textgreater}\\
single integer only, for example 3\\
\textbf{\textless /answer\textgreater}

\end{tcolorbox}
\caption{Shared system prompt used in both SFT and GRPO. It defines the role, PTC persona, reasoning guide, question, and required output format.}
\label{fig:system-prompt-template}
\end{figure}

\noindent Figure~\ref{fig:method-overview} provides an overview of the proposed framework for simulating tenant responses to energy policy interventions. The framework has two connected components. First, each survey instance is converted into a structured prompt with three layers. A system instruction asks the model to answer as one specific resident and remain consistent with that person's housing situation, attitudes, and priorities. A PTC persona prompt then adds barrier- or benefit-oriented persona descriptions, such as \textit{Financial Sensitive} and \textit{Immediate Utility Seekers}, to capture the frictions and motivations most relevant to the intervention. A PTC reasoning prompt further asks the model to reflect briefly on burden, uncertainty, personal gains, and the likely direction of the response before answering the survey question in the required format. Second, we compare different modeling strategies for generating these responses. In addition to a prompt-only GPT-3.5-turbo baseline, we adapt local open-weight models using QLoRA under two training settings: supervised fine-tuning (SFT), which learns directly from survey examples, and Group Relative Policy Optimization (GRPO), which further optimizes outputs with an explicit reward. The following subsections explain how PTC personas are designed and assigned, how the shared prompt is structured, and how SFT and GRPO are used to adapt the local models.

\subsection{Prompt Design and PTC Persona Construction}

\noindent As shown in Figure~\ref{fig:system-prompt-template}, each survey prompt combines five components: a system instruction, PTC persona descriptions, a PTC reasoning checklist, the survey question, and an output format template. The system instruction and question format provide structural scaffolding common to all instances, while the persona and reasoning components are the substantive PTC-theoretic elements that vary by respondent. The following subsubsections introduce the two core components in detail, followed by a brief note on the remaining structural elements.

\subsubsection{PTC Persona Prompting}

We adopt the EER persona setup developed in prior work on Dutch social-housing tenants \cite{anonymous2025beyond}. In that study, tenant personas are obtained from survey indicators of perceived renovation benefits and barriers. In our study, we adopt the persona predictor developed in that work, which is built to infer the most likely barrier and benefit persona from non-EER contextual variables, such as age, household composition, tenure length, and other socio-demographic characteristics. \textit{Benefit personas} (BE1--BE5) measure how strongly tenants value outcomes such as comfort, health, wellbeing, reduced energy consumption, environmental improvement, and neighborhood effects. \textit{Barrier personas} (BA1--BA7) measure perceived frictions that may prevent agreement with renovation plans, including rent or service-charge increases, temporary relocation, daily-life disruption, lack of time or information, distrust, and uncertainty. 


In our LLM framework, each survey instance is assigned one \textit{benefit persona} and one \textit{barrier persona} according to the respondent's observed survey pattern and the persona predictors developed in the prior study. The \textit{benefit persona} captures what kinds of EER gains the respondent is likely to value, while the \textit{barrier persona} captures the transaction-cost frictions and behavioral constraints that may shape the respondent's answer. This keeps persona construction tied to empirical survey evidence rather than ad hoc role descriptions or demographic stereotypes. Table~\ref{tab:persona-summary} summarizes the twelve persona types; full definitions appear in the Appendix.

\begin{table}[t]
\centering
\small
\resizebox{\columnwidth}{!}{%
\begin{tabular}{|l|c|l|}
\hline
\textbf{Persona type} & \textbf{Index} & \textbf{Personas} \\
\hline
\multirow{7}{*}{Barrier} & BA1 & \textit{Financial Sensitive} \\
 & BA2 & \textit{Practical and Financial Concerned} \\
 & BA3 & \textit{Ambivalent Observer} \\
 & BA4 & \textit{Self-Reliant but Unconvinced} \\
 & BA5 & \textit{Financially Alarmed} \\
 & BA6 & \textit{Confident Acceptors} \\
 & BA7 & \textit{The Uncertain} \\
\hline
\multirow{5}{*}{Benefit} & BE1 & \textit{Immediate Utility Seekers} \\
 & BE2 & \textit{Personal Comfort Seekers} \\
 & BE3 & \textit{Balanced Benefit Idealists} \\
 & BE4 & \textit{Ambivalent Respondents} \\
 & BE5 & \textit{Pessimists} \\
\hline
\end{tabular}%
}
\caption{PTC-aware personas used for prompt conditioning. \textit{Barrier personas} (BA1--BA7) encode frictions that may impede renovation agreement; \textit{benefit personas} (BE1--BE5) encode values that may motivate it. Each respondent is assigned one barrier and one benefit persona. Full definitions are in the Appendix.}
\label{tab:persona-summary}
\end{table}

\subsubsection{PTC Reasoning Prompting}

The reasoning component asks the model to briefly consider four items before producing a final Likert-scale answer: the burden the intervention imposes, sources of uncertainty, expected personal gains, and the likely direction of the response. This design anchors outputs to the explicit decision frictions encoded in the assigned PTC persona, rather than relying on demographic priors alone. Figure~\ref{fig:answer-example} illustrates the expected output format, where a short reasoning trace enclosed in \texttt{<thinking>} tags precedes a single integer answer.

\begin{figure}[!t]
\centering
\begin{tcolorbox}[
    colback=green!12,
    colframe=green!45!black,
    arc=2mm,
    boxrule=0.5pt,
    left=2mm,
    right=2mm,
    top=2mm,
    bottom=2mm,
    width=0.98\linewidth
]

\footnotesize

\noindent \textbf{\textless thinking\textgreater}\\
Thermal comfort matters to me because it improves daily life right away.
I am also careful about money, so I would want to know that renovation will not raise my rent or service charges too much.
If the comfort gains are clear and the financial risk is controlled, I would likely care about this benefit.\\
\textbf{\textless /thinking\textgreater}

\vspace{0.3em}

\noindent \textbf{\textless answer\textgreater}\\
4\\
\textbf{\textless /answer\textgreater}

\end{tcolorbox}
\caption{Formatted answer example corresponding to the shared survey prompt.}
\label{fig:answer-example}
\end{figure}

\subsubsection{System Prompt and Survey Questions}

The system instruction briefly positions the model as a survey respondent with the assigned persona characteristics, ensuring consistent role adoption throughout the response. The survey question is drawn verbatim from the original EER questionnaire and appended after the persona and reasoning prompts. The output format template, shown in Figure~\ref{fig:system-prompt-template}, specifies the required \texttt{<thinking>} and \texttt{<answer>} tag structure so that training targets have a consistent form across both SFT and GRPO.

\subsection{Supervised Fine-Tuning}
\noindent Supervised fine-tuning adapts the base model by maximizing the likelihood of survey-consistent responses conditioned on the prompt. Given a dataset \(D=\{(x_i,y_i)\}_{i=1}^N\), where \(x_i\) is the persona-intervention prompt and \(y_i\) is the target response, SFT minimizes the negative log likelihood
\begin{equation}
\begin{aligned}
\mathcal{L}_{\mathrm{SFT}}(\theta)
&= - \sum_{(x,y)\in D}
\sum_{t=1}^{|y|}
\log \pi_\theta(y_t \mid x,y_{<t}).
\end{aligned}
\end{equation}
This objective is appropriate when the goal is to teach the model the format, domain vocabulary, and empirical mapping between survey-conditioned prompts and observed answers. SFT also provides a transparent baseline because it uses only labeled demonstrations and does not require a separately specified reward function \cite{ouyang2022training}. In our implementation, SFT is QLoRA-based: the base model remains quantized and fixed, while low-rank adapter parameters are trained to reproduce the survey-consistent reasoning trace and final integer answer \cite{hu2021lora,dettmers2023qlora}.

\subsection{Group Relative Policy Optimization}
\noindent GRPO is a reinforcement-learning method introduced to improve LLM reasoning while reducing the memory overhead of critic-based PPO \cite{shao2024deepseekmath}. As with SFT, the GRPO run uses QLoRA adapters rather than full-parameter updates. For each prompt \(q\), the current policy samples a group of \(G\) candidate responses \(\{o_1,\ldots,o_G\}\). The reward function assigns each response a scalar reward \(r_i\), and the reward is normalized within the sampled group to obtain a token-level advantage:
\begin{equation}
\begin{aligned}
\hat{A}_{i,t}
&= \frac{r_i-\mu}{\sigma}, \\
\mu
&= \frac{1}{G}\sum_{i=1}^{G} r_i, \quad
\sigma =
\sqrt{\frac{1}{G}\sum_{i=1}^{G}(r_i-\mu)^2+\epsilon}.
\end{aligned}
\end{equation}
The policy is then updated with a GRPO objective and a KL penalty that keeps the adapted model close to a reference policy:
\begin{equation}
\begin{aligned}
p_{i,t}^{\theta}
&= \pi_\theta(o_{i,t}\mid q,o_{i,<t}), \\
p_{i,t}^{\mathrm{ref}}
&= \pi_{\mathrm{ref}}(o_{i,t}\mid q,o_{i,<t}).
\end{aligned}
\end{equation}
\begin{equation}
\begin{aligned}
\mathcal{L}_{\mathrm{GRPO}}(\theta)
&= -\frac{1}{\sum_{i=1}^{G}|o_i|}
\sum_{i=1}^{G}\sum_{t=1}^{|o_i|}
\Biggl[
\frac{p_{i,t}^{\theta}}{[p_{i,t}^{\theta}]_{\mathrm{sg}}}
\hat{A}_{i,t} \\
&\quad
- \beta_{\mathrm{KL}}
D_{\mathrm{KL}}(\pi_\theta \| \pi_{\mathrm{ref}})
\Biggr],
\end{aligned}
\end{equation}
where \([\cdot]_{\mathrm{sg}}\) denotes stop-gradient. The KL term is estimated at each generated token as
\begin{equation}
\begin{aligned}
D_{\mathrm{KL}}(\pi_\theta \| \pi_{\mathrm{ref}})
&= \frac{p_{i,t}^{\mathrm{ref}}}{p_{i,t}^{\theta}}
- \log
\frac{p_{i,t}^{\mathrm{ref}}}{p_{i,t}^{\theta}}
-1.
\end{aligned}
\end{equation}
Unlike actor-critic PPO, GRPO does not require a learned value model; the group baseline provides the relative advantage estimate. This makes it attractive for resource-constrained fine-tuning of open models while still allowing explicit optimization of response quality.

In this paper, GRPO is used to compare whether reward-guided adaptation improves policy-response simulation over SFT alone. The reward combines survey-label accuracy, output-format validity, and length control:
\begin{equation}
\begin{aligned}
r_i
&= w_1 r_{\mathrm{acc},i}
+ w_2 r_{\mathrm{format},i}
+ w_3 r_{\mathrm{len},i}.
\end{aligned}
\end{equation}
The accuracy reward decreases smoothly as the predicted integer answer \(\hat{y}_i\) moves away from the observed survey label \(y_i\):
\begin{equation}
\begin{aligned}
r_{\mathrm{acc},i}
&= r_{\min}
+ \frac{r_{\max}-r_{\min}}{1+e_i^2},
\quad
e_i = |\hat{y}_i-y_i|.
\end{aligned}
\end{equation}
The format reward checks whether the model returns a parsable reasoning field and a final single-integer answer field:
\begin{equation}
r_{\mathrm{format},i} =
\begin{cases}
1, & \text{if the required format is correct},\\
0, & \text{otherwise}.
\end{cases}
\end{equation}
In implementation, this check can be applied to XML-like tags such as
\(\langle\mathrm{thinking}\rangle\cdots\langle/\mathrm{thinking}\rangle\)
and
\(\langle\mathrm{answer}\rangle\cdots\langle/\mathrm{answer}\rangle\),
or to the equivalent \emph{Thinking}/\emph{Answer} fields in the shared prompt. Finally, the length reward penalizes reasoning traces and answer fields that deviate from the target lengths:
\begin{equation}
\begin{aligned}
r_{\mathrm{len},i}
&= -\Bigl(
\alpha_\ell |L_{r,i}-L_r^{\mathrm{target}}| \\
&\quad
+ \beta_\ell |L_{a,i}-L_a^{\mathrm{target}}|
\Bigr),
\end{aligned}
\end{equation}
where \(L_{r,i}\) and \(L_{a,i}\) are the reasoning and answer lengths. This reward design favors outputs that are accurate with respect to the survey label, short enough to remain inspectable, and strict enough to be parsed automatically.

\section{\uppercase{Experiments and Results}}
\subsection{Evaluation Metrics}
\noindent We evaluate the models at the individual-response level using two complementary metrics. For ordinal prediction error, we use mean absolute error (MAE) between the model's integer response \(\hat{y}_i\) and the observed survey label \(y_i\):
\begin{equation}
\mathrm{MAE}
= \frac{1}{N}\sum_{i=1}^{N}\left|\hat{y}_i-y_i\right|.
\end{equation}
Lower MAE indicates closer agreement with the observed survey labels.

For classification performance, we use two accuracy metrics. Exact-match accuracy ($\mathrm{Acc}$) counts only predictions that equal the true label:
\begin{equation}
\mathrm{Acc}
= \frac{1}{N}\sum_{i=1}^{N}\mathbf{1}[\hat{y}_i=y_i].
\end{equation}
Relaxed accuracy ($\mathrm{Acc{\pm}1}$) counts predictions within one scale point of the true label, capturing near-miss agreement on the ordinal response scale:
\begin{equation}
\mathrm{Acc{\pm}1}
= \frac{1}{N}\sum_{i=1}^{N}\mathbf{1}[|\hat{y}_i-y_i|\leq 1].
\end{equation}
Together, MAE, Acc, and Acc$\pm$1 distinguish exact classification, near-miss agreement, and overall ordinal distance from the observed survey responses.

\begin{figure}
\centering
\includegraphics[width=0.48\textwidth]{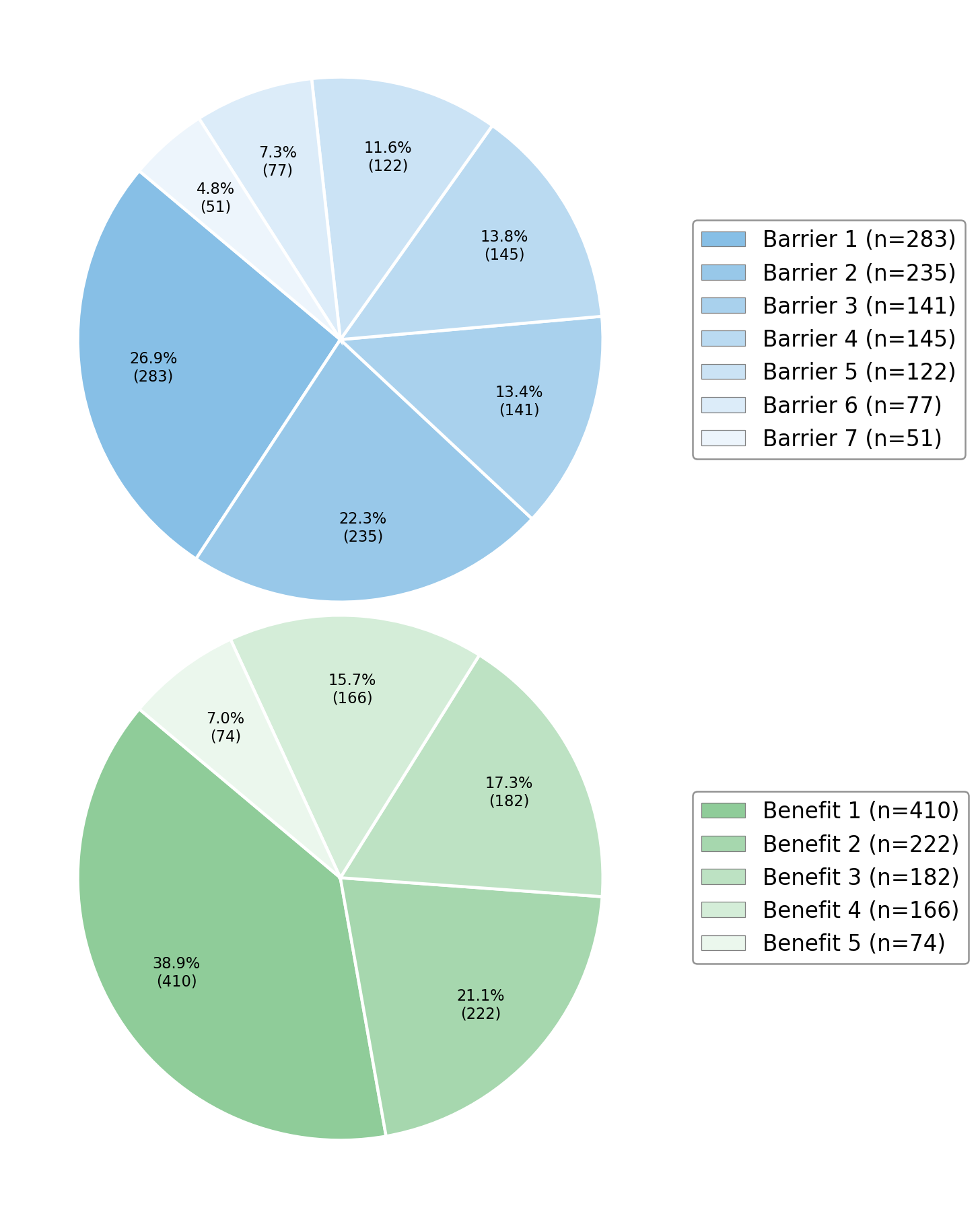}
\caption{Distribution of the twelve PTC personas in the dataset.}
\label{fig:persona-distribution}
\end{figure}

\subsection{Experimental Setup}
\noindent The dataset contains approximately 1,068 tenant consultation records and around 40,548 question--answer pairs collected from real respondents. Of these pairs, 24,242 are attitude-related and the remainder are demographic. For fine-tuning, we use 4,600 attitude-related question--answer pairs, corresponding to around 20\% of the attitude-related subset, and evaluate the models on the remaining attitude-related pairs. Each attitude-related pair captures a tenant's response to an energy policy item, including benefit- and barrier-related questions answered on a 5-point Likert scale. Figure~\ref{fig:persona-distribution} shows the distribution of the twelve PTC personas in the dataset. In addition to survey responses, each record includes basic socio-demographic information, such as age, income or salary level, and household composition (e.g., family size and presence of dependants). These attributes support the construction of the PTC persona used during prompting and fine-tuning.

We use GPT-3.5-turbo as a prompt-only baseline under three prompting conditions that progressively add PTC information. The first condition uses no PTC persona or reasoning prompt; the second adds the PTC persona prompt; and the third adds both the PTC persona prompt and the PTC reasoning prompt. For the local open-weight models, we use Ministral-8B-Instruct and Llama-3.1-8B-Instruct, both served via Ollama. These models are first evaluated without fine-tuning under the fully enriched prompt, which includes both persona and reasoning components, and are then adapted with QLoRA-based SFT and QLoRA-based GRPO. Table~\ref{tab:model-configs} summarizes the nine experimental conditions. The GPT-3.5-turbo conditions (a--c) isolate the contribution of PTC prompting, while the open-weight model conditions (-1 through -3) keep the fully enriched prompt fixed and vary only the adaptation method.

Both open-weight models are adapted using QLoRA with 4-bit NF4 quantization, double quantization, and fp16 compute precision. For SFT, we attach LoRA adapters ($r{=}16$, $\alpha{=}32$, dropout 0.05) to all linear layers, use a learning rate of $1\times10^{-6}$ and an effective batch size of 8, and supervise only the final response token. For GRPO, we use the same QLoRA configuration and learning rate, with a per-device batch size of 4, 4 completions sampled per prompt, gradient clipping at 0.1, and bf16 mixed precision.

\begin{figure*}[t]
\centering
\includegraphics[width=\textwidth]{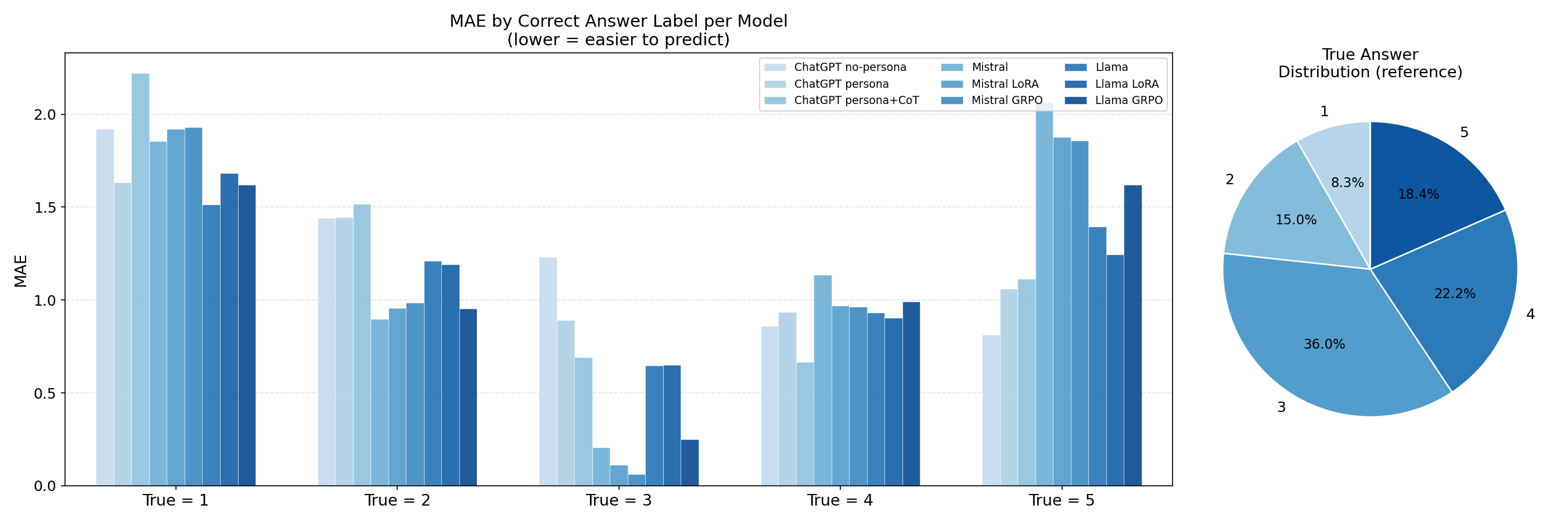}
\caption{MAE by true answer label for each model condition. Lower values indicate that the corresponding response category is easier to predict.}
\label{fig:answer-distribution}
\end{figure*}

\begin{figure*}[htp]
\centering
\includegraphics[width=\textwidth]{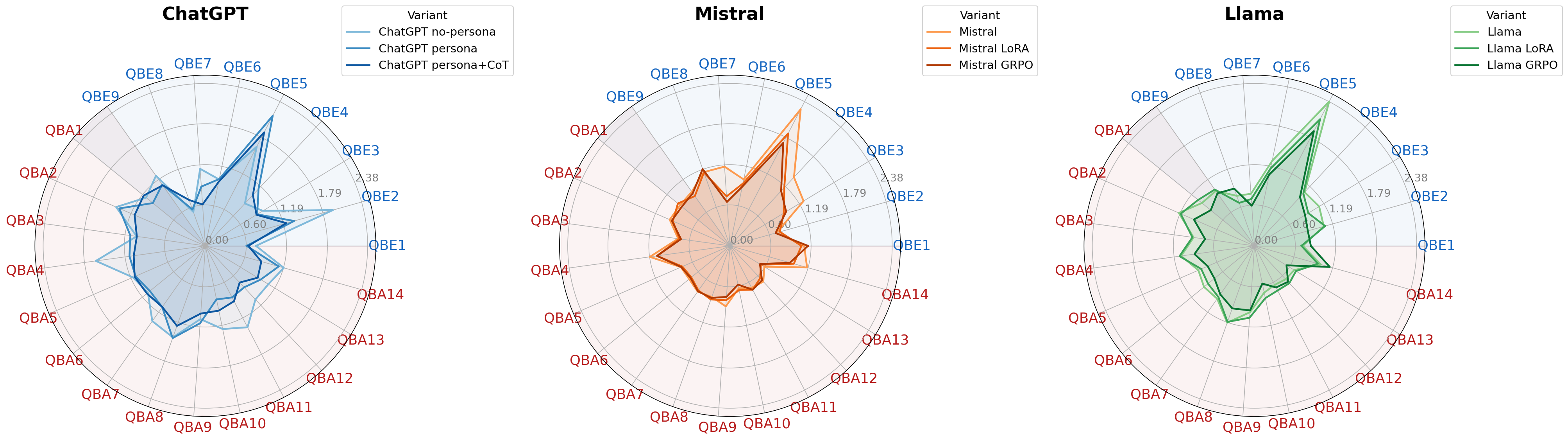}
\caption{Question-wise MAE across all model conditions. QBE and QBA index the benefit- and barrier-related survey questions. Note: these question indices are distinct from the PTC persona labels (BE1--BE5, BA1--BA7) in Table~\ref{tab:persona-summary}.}
\label{fig:question-wise-model}
\end{figure*}

\begin{table}[t]
\centering
\setlength{\tabcolsep}{3pt}
\footnotesize
\begin{tabular}{lccl}
\hline
\textbf{Model} & \textbf{PTC Per.} & \textbf{PTC Reas.} & \textbf{Fine-tune} \\
\hline
GPT-3.5-turbo-a    & No  & No  & ---  \\
GPT-3.5-turbo-b    & Yes & No  & ---  \\
GPT-3.5-turbo-c    & Yes & Yes & ---  \\
\hline
Ministral-8B-Instruct-1 & Yes & Yes & None \\
Ministral-8B-Instruct-2 & Yes & Yes & SFT  \\
Ministral-8B-Instruct-3 & Yes & Yes & GRPO \\
\hline
Llama-3.1-8B-Instruct-1 & Yes & Yes & None \\
Llama-3.1-8B-Instruct-2 & Yes & Yes & SFT  \\
Llama-3.1-8B-Instruct-3 & Yes & Yes & GRPO \\
\hline
\end{tabular}
\caption{Experimental conditions. PTC~Per./Reas.\ = PTC persona/reasoning prompting; open-weight models served via Ollama, fine-tuned with QLoRA.}
\label{tab:model-configs}
\end{table}

\subsection{Main Results}
\noindent Table~\ref{tab:benchmark-results} presents the full benchmark results. The three GPT-3.5-turbo conditions show that PTC-based prompting improves performance step by step. The baseline without persona or reasoning prompts (GPT-3.5-turbo-a) performs worst overall (MAE~1.4783, Acc~0.2476, Acc$\pm$1~0.6780). Adding a PTC-based persona prompt (GPT-3.5-turbo-b) substantially improves performance, reducing MAE to 1.0859 and raising Acc to 0.3131 and Acc$\pm$1 to 0.7168. Adding PTC-based reasoning (GPT-3.5-turbo-c) yields a further gain in ordinal fit, with the lowest MAE (1.0335) and highest Acc$\pm$1 (0.7379) among the three GPT-3.5-turbo settings, although exact-match accuracy remains slightly below GPT-3.5-turbo-b (0.3086 versus 0.3131). Taken together, these results show that PTC-based persona and reasoning prompts consistently improve simulation quality, supporting the effectiveness of the proposed PTC-based modeling strategy.

Table~\ref{tab:benchmark-results} also shows that fine-tuning local open-weight models further outperforms the prompt-only baseline. Compared with the best GPT-3.5-turbo prompt-only setting (GPT-3.5-turbo-c: MAE~1.0335, Acc~0.3086, Acc$\pm$1~0.7379), both SFT and GRPO achieve lower MAE and higher accuracy on Ministral-8B-Instruct and Llama-3.1-8B-Instruct. The strongest overall result comes from Llama-3.1-8B-Instruct with GRPO, which achieves the lowest MAE (0.8694) and highest Acc$\pm$1 (0.7843), while Ministral-8B-Instruct with SFT attains the highest exact-match accuracy in the table (0.3702). For both model families, GRPO yields the best MAE and Acc$\pm$1, suggesting that fine-tuning improves not only exact matches but also the ordinal closeness of predictions to the observed survey responses.

\begin{table}[t]
\centering
\small
\setlength{\tabcolsep}{4pt}
\resizebox{\columnwidth}{!}{%
\begin{tabular}{lrrr}
\hline
Experiment & MAE & Acc & Acc$\pm$1 \\
\hline
GPT-3.5-turbo-a & 1.4783 & 0.2476 & 0.6780 \\
GPT-3.5-turbo-b & 1.0859 & \textbf{0.3131} & 0.7168 \\
GPT-3.5-turbo-c & \textbf{1.0335} & 0.3086 & \textbf{0.7379} \\
\hline
Ministral-8B-Instruct-1 & 0.9855 & 0.3466 & 0.7338 \\
Ministral-8B-Instruct-2 & 0.8962 & \textbf{0.3702} & 0.7608 \\
Ministral-8B-Instruct-3 & \textbf{0.8802} & 0.3683 & \textbf{0.7665} \\
\hline
Llama-3.1-8B-Instruct-1 & 1.0012 & 0.3320 & 0.7482 \\
Llama-3.1-8B-Instruct-2 & 0.9780 & 0.3383 & 0.7530 \\
Llama-3.1-8B-Instruct-3 & \textbf{0.8694} & \textbf{0.3643} & \textbf{0.7843} \\
\hline
\end{tabular}%
}
\caption{Benchmark results across prompt-only and fine-tuned settings. Lower MAE and higher Acc/Acc$\pm$1 indicate better agreement with observed survey responses.}
\label{tab:benchmark-results}
\end{table}

\begin{table*}[t]
\centering
\scriptsize
\setlength{\tabcolsep}{4pt}
\begin{tabular}{lrrrrrrr}
\hline
\textbf{Model} & \textbf{Fin.\ Sensitive} & \textbf{Prac.\ \& Fin.\ Concerned} & \textbf{Ambivalent Obs.} & \textbf{Self-Reliant} & \textbf{Fin.\ Alarmed} & \textbf{Confident Accept.} & \textbf{The Uncertain} \\
\hline
GPT-3.5-turbo-a    & \maeheat{1.1587} & \maeheat{1.0640} & \maeheat{1.2909} & \maeheat{1.1806} & \maeheat{1.1410} & \maeheat{1.5000} & \maeheat{1.1172} \\
GPT-3.5-turbo-b    & \maeheat{1.2315} & \maeheat{1.1014} & \maeheat{0.8071} & \maeheat{1.1330} & \maeheat{1.1801} & \maeheatbf{0.8348} & \maeheat{0.9752} \\
GPT-3.5-turbo-c    & \maeheat{1.2029} & \maeheat{0.8750} & \maeheat{0.9476} & \maeheat{0.8624} & \maeheatbf{1.0651} & \maeheat{1.2007} & \maeheat{0.7268} \\
\hline
Ministral-8B-Instruct-1 & \maeheat{1.1543} & \maeheat{1.0110} & \maeheat{0.7468} & \maeheat{0.6672} & \maeheat{1.4686} & \maeheat{1.4222} & \maeheat{0.5052} \\
Ministral-8B-Instruct-2 & \maeheat{1.0358} & \maeheatbf{0.9088} & \maeheatbf{0.7117} & \maeheat{0.6029} & \maeheat{1.2942} & \maeheat{1.3607} & \maeheat{0.4526} \\
Ministral-8B-Instruct-3 & \maeheat{1.0007} & \maeheat{0.9156} & \maeheat{0.7210} & \maeheatbf{0.5978} & \maeheat{1.2255} & \maeheat{1.3779} & \maeheatbf{0.4037} \\
\hline
Llama-3.1-8B-Instruct-1 & \maeheat{1.1248} & \maeheat{0.9540} & \maeheat{0.8819} & \maeheat{0.9550} & \maeheat{1.1601} & \maeheat{0.9530} & \maeheat{0.8577} \\
Llama-3.1-8B-Instruct-2 & \maeheat{1.0930} & \maeheat{0.9634} & \maeheat{0.8698} & \maeheat{0.8896} & \maeheat{1.1505} & \maeheat{0.8949} & \maeheat{0.8643} \\
Llama-3.1-8B-Instruct-3 & \maeheatbf{0.9970} & \maeheat{0.9326} & \maeheat{0.7233} & \maeheat{0.6442} & \maeheat{1.1397} & \maeheat{1.1186} & \maeheat{0.4632} \\
\hline
\textbf{Avg (all)} & \maeheat{1.1110} & \maeheat{0.9695} & \maeheat{0.8556} & \maeheat{0.8370} & \maeheat{1.2028} & \maeheat{1.1847} & \maeheat{0.7073} \\
\hline
\end{tabular}
\caption{MAE [-] by \textit{barrier persona} type (see Table~\ref{tab:persona-summary} for persona definitions).}
\label{tab:mae-barrier}
\end{table*}

\begin{table*}[t]
\centering
\scriptsize
\setlength{\tabcolsep}{4pt}
\begin{tabular}{lrrrrr}
\hline
\textbf{Model} & \textbf{Immed.\ Utility Seek.} & \textbf{Personal Comfort Seek.} & \textbf{Balanced Ideal.} & \textbf{Ambivalent Resp.} & \textbf{Pessimists} \\
\hline
GPT-3.5-turbo-a    & \maeheat{1.1223} & \maeheat{1.2352} & \maeheat{1.1031} & \maeheat{1.2551} & \maeheat{1.4786} \\
GPT-3.5-turbo-b    & \maeheat{1.1876} & \maeheat{1.3731} & \maeheatbf{0.8910} & \maeheat{0.7449} & \maeheat{1.2857} \\
GPT-3.5-turbo-c    & \maeheat{1.0125} & \maeheat{1.1507} & \maeheat{1.0828} & \maeheat{0.7868} & \maeheat{1.1411} \\
\hline
Ministral-8B-Instruct-1 & \maeheat{0.9500} & \maeheat{1.2332} & \maeheat{1.2671} & \maeheat{0.6183} & \maeheat{0.9169} \\
Ministral-8B-Instruct-2 & \maeheat{0.8234} & \maeheat{1.1083} & \maeheat{1.2353} & \maeheat{0.5827} & \maeheat{0.8400} \\
Ministral-8B-Instruct-3 & \maeheatbf{0.7838} & \maeheat{1.0787} & \maeheat{1.2791} & \maeheatbf{0.5743} & \maeheatbf{0.8340} \\
\hline
Llama-3.1-8B-Instruct-1 & \maeheat{1.0355} & \maeheat{1.2267} & \maeheat{0.9234} & \maeheat{0.7028} & \maeheat{1.2984} \\
Llama-3.1-8B-Instruct-2 & \maeheat{1.0171} & \maeheat{1.1845} & \maeheat{0.9109} & \maeheat{0.7052} & \maeheat{1.1858} \\
Llama-3.1-8B-Instruct-3 & \maeheat{0.8239} & \maeheatbf{1.0364} & \maeheat{1.0713} & \maeheat{0.6072} & \maeheat{0.9894} \\
\hline
\textbf{Avg (all)} & \maeheat{0.9729} & \maeheat{1.1807} & \maeheat{1.0849} & \maeheat{0.7308} & \maeheat{1.1078} \\
\hline
\end{tabular}
\caption{MAE [-] by \textit{benefit persona} type (see Table~\ref{tab:persona-summary} for persona definitions). Model labels follow Table~\ref{tab:model-configs}. Bold = lowest MAE per column.}
\label{tab:mae-benefit}
\end{table*}

\subsection{Question-Wise Analysis}
\noindent Figure~\ref{fig:answer-distribution} shows the MAE by true answers for each model. Middle response categories are generally easier to predict than extreme ones. Across model conditions, MAE is lowest for label~3 and remains relatively low for labels~2 and~4, whereas labels~1 and~5 are consistently harder. The pie chart suggests one reason for this pattern: the extreme categories are also less common in the data, with label~1 accounting for only 8.3\% of responses and label~5 for 18.4\%, compared with 36.0\% for label~3 and 22.2\% for label~4. In other words, the rarest response categories are also the most difficult to model accurately. Figure~\ref{fig:answer-distribution} also helps explain the differences in Table~\ref{tab:benchmark-results}: PTC-based prompting and model adaptation often reduce ordinal error by moving predictions closer to the correct label, even when exact-match accuracy improves only modestly.

Figure~\ref{fig:question-wise-model} plots question-wise MAE across all model conditions, allowing comparison of how prompting and fine-tuning affect performance on individual benefit and barrier items. The figure shows that these gains are not uniform across questions, but two broad patterns are clear. For GPT-3.5-turbo, the improvement comes from prompt design: adding a PTC-based persona generally improves the baseline across many items, and adding PTC-based reasoning often yields further but smaller gains. For the open-weight models, the main improvement comes from model adaptation. Both Ministral-8B-Instruct and Llama-3.1-8B-Instruct show lower question-wise error after SFT and GRPO than in their untuned settings, indicating that fine-tuning provides a more consistent benefit across the full question set. One notable exception is BE5, which remains difficult across all model families and settings.

\subsection{Persona-Wise Analysis}

\noindent Tables~\ref{tab:mae-barrier} and~\ref{tab:mae-benefit} report MAE disaggregated by \textit{barrier persona} and \textit{benefit persona} type, where each cell shows the average prediction error for survey instances assigned to that persona. Three patterns emerge that complement the aggregate results in Table~\ref{tab:benchmark-results}.

First, persona difficulty is uneven and follows a recognizable structure. Among \textit{barrier personas}, \textit{Financially Alarmed} (avg.\ MAE 1.20) and \textit{Confident Acceptors} (avg.\ 1.18) are consistently the hardest across all models, while \textit{The Uncertain} (avg.\ 0.71) and \textit{Self-Reliant but Unconvinced} (avg.\ 0.84) are the easiest. Among \textit{benefit personas}, \textit{Personal Comfort Seekers} (avg.\ 1.18) and \textit{Balanced Benefit Idealists} (avg.\ 1.08) are hardest, while \textit{Ambivalent Respondents} (avg.\ 0.73) is easiest. The easy personas in both groups represent moderate or undecided tenants who tend to cluster near the neutral middle of the response scale, making them easier to predict.

Second, fine-tuning with GRPO generally dominates but not universally. Ministral-8B-Instruct-3 wins 4 of 7 barrier columns and 3 of 5 benefit columns. However, two notable exceptions arise: \textit{Confident Acceptors} (barrier) and \textit{Balanced Benefit Idealists} (benefit) are both best predicted by GPT-3.5-turbo-b, the persona-only prompt without reasoning or fine-tuning. This suggests that for polarized-positive tenant profiles, PTC persona prompting alone captures sufficient context, while reward-guided adaptation does not provide additional benefit.

Third, Ministral and Llama models exhibit different persona-level weaknesses. Ministral models consistently struggle on \textit{Confident Acceptors} and \textit{Financially Alarmed} (MAE above 1.2 even after fine-tuning), whereas Llama models are more moderate on those columns. Conversely, Llama-3.1-8B-Instruct-3 outperforms Ministral-8B-Instruct-3 on \textit{Personal Comfort Seekers} (1.04 versus 1.08). Finally, all Ministral fine-tuned models perform worse than GPT-3.5-turbo-a on \textit{Balanced Benefit Idealists}, suggesting that fine-tuning may introduce bias for this holistic persona type.

\section{\uppercase{Conclusion}}
\label{sec:conclusion}

\noindent This paper presented a framework for LLM-based simulation of tenant responses to EER interventions, grounded in PTC theory. Rather than using generic role descriptions or demographic profiles, the framework conditions each simulation instance on empirically derived PTC personas from prior survey research on Dutch social-housing tenants. These personas encode the specific frictions and motivations shaping a respondent's evaluation of a renovation policy, and are paired with a structured PTC reasoning prompt that guides the model to deliberate over burden, uncertainty, and personal gain before answering. This design makes transaction costs an explicit part of the model's deliberative context, rather than a post-hoc interpretation of its outputs.

Across the experiments, the results support three main conclusions. First, PTC-aware prompting improves simulation quality even in a prompt-only setting: for GPT-3.5-turbo, adding persona information and reasoning substantially reduces MAE and improves accuracy relative to a no-persona baseline. Second, adapting local open-weight models yields stronger overall performance than prompt-only baselines, with the best overall results achieved by Llama-3.1-8B-Instruct under GRPO. Third, the choice of metric matters: GRPO most consistently improves ordinal closeness, while exact-match accuracy can show smaller or more model-specific gains.

The finer-grained analyses also show that model performance is uneven across response types. Moderate responses are easier to predict than extreme responses, and improvements are distributed differently across questions and model families. These findings suggest that LLM-based policy simulation should be evaluated not only by aggregate averages, but also by response distribution and question-level behavior.

Overall, the results suggest that PTC-based persona design is a useful bridge between institutional policy theory and LLM agent modeling. It improves interpretability, supports inspectable adaptation of open-weight models, and provides a practical direction for building more policy-relevant simulations of tenant behavior.

\section*{\uppercase{Acknowledgements}}
\begin{itemize}
  \item \textbf{Conflicts of Interest:} 
  The authors have no conflicts of interest or competing interests to declare in the context of this paper.
  \item \textbf{Financial and other Support:} This research was supported by the Align4Energy Project (NWA.1389.20.251) and utilized the Dutch National e-Infrastructure with the support from the SURF Cooperative (grant number: EINN-5398).
  \item \textbf{Author Contributions:}  Weijie Xia contributed to conceptualization, implementation, experiments, and writing of the manuscript. Stefanie Horian contributed to conceptualization and writing. Hanyue Huang contributed to implementation. Queena K. Qian and Jie Yang contributed to the revision. Pedro P. Vergara contributed to funding acquisition and revision.
  \item \textbf{Data Sharing:} Code and data are available at \href{https://github.com/xiaweijie1996/socialagent}{Personal Repo} and \href{https://github.com/distributionnetworksTUDelft/LLMAgentEnergyCitizent}{TU Delft Repo}.
  \item \textbf{AI Tools:} 
  The paper used AI tools for text correction purposes only.
\end{itemize}

\bibliographystyle{apalike}
{\small
\bibliography{aaai2026}}

\section*{\uppercase{Appendix}}
\subsection{PTC Persona Definitions}
\label{sec:persona-definitions}
The prompt uses two persona fields: a \textit{barrier persona} that captures the respondent's perceived obstacles to renovation, and a \textit{benefit persona} that captures the respondent's perceived value of renovation. These personas are used as compact PTC profiles in the shared survey prompt.

\begin{description}[leftmargin=0pt, itemsep=2pt, topsep=2pt, parsep=0pt, font=\normalfont\bfseries]
\item[Financial Sensitive.] This persona is generally supportive of renovation but highly sensitive to financial consequences, especially potential increases in rent or service charges. Practical disruptions are less concerning, but the respondent needs strong reassurance that renovation will not worsen their financial situation.

\item[Practical and Financial Concerned.] This persona perceives both financial and practical barriers as major obstacles. The respondent worries about temporary relocation, daily-life disruption, nuisance, and long renovation timelines, and may also feel overloaded or lack time and clear information. Step-by-step planning and logistical support are essential.

\item[Ambivalent Observer.] This persona feels mostly neutral about renovation barriers and tends to observe rather than actively engage. The respondent is not strongly opposed, but is not proactive either. They prefer minimal involvement and respond better to simple visuals or relatable messages than to detailed explanations.

\item[Self-Reliant but Unconvinced.] This persona often feels uncertain about both financial and practical aspects of renovation and may frequently think, ``I don't know.'' The respondent trusts their own judgment but lacks a clear understanding of what renovation would mean for them. Personalized explanations, such as home visits, are more effective than generic campaigns.

\item[Financially Alarmed.] This persona is extremely worried about the financial impact of renovation. The respondent strongly believes it will increase rent, energy bills, and service charges. Practical disruptions such as relocation or delays intensify these concerns. Without firm financial guarantees, they are very likely to reject renovation.

\item[Confident Acceptors.] This persona perceives very few barriers to renovation. The respondent feels financially secure, trusts the process, and is willing to tolerate temporary inconvenience. They are often already familiar with alternative heating systems and could serve as a positive reference point for others.

\item[The Uncertain.] This persona frequently feels unsure when thinking about renovation and often responds with ``I don't know.'' This reflects limited information or difficulty processing the topic rather than active opposition. The respondent benefits most from guided, personal support through trusted intermediaries or in-home assistance.

\item[Immediate Utility Seekers.] This persona cares strongly about immediate and tangible benefits from renovation, such as better indoor comfort, improved health, personal wellbeing, and lower energy consumption. Aesthetic upgrades or neighborhood-level improvements are less important. The respondent is generally positive about energy-efficiency renovation when short-term personal gains are explained clearly and concretely.

\item[Personal Comfort Seekers.] This persona mainly cares about how renovation affects personal comfort and health. Environmental benefits, building appearance, and neighborhood improvements matter little. The respondent tends to rely on trust-based reassurance and prefers discussion events or face-to-face explanations over purely digital communication.

\item[Balanced Benefit Idealists.] This persona values a broad range of renovation benefits. Personal comfort is important, but so are environmental responsibility and positive impacts on the community. The respondent responds well to holistic narratives that connect personal wellbeing with social and environmental goals, and appreciates participatory or interactive engagement formats.

\item[Ambivalent Respondents.] This persona feels mostly neutral about the benefits of renovation. The topic does not feel particularly relevant or urgent, and the attitude reflects uncertainty rather than clear support or opposition. The respondent prefers simple, low-effort communication that clarifies why renovation should matter personally.

\item[Pessimists.] This persona generally does not care much about the proposed benefits of renovation. Most advantages feel irrelevant, although reducing energy consumption has some limited appeal. The respondent tends to avoid engagement, including emails or events, and benefit-focused messaging alone is unlikely to motivate them.
\end{description}

\end{document}